\documentclass[review]{elsarticle}

\usepackage{algorithm} 
\usepackage{etoolbox,lipsum}
\usepackage{amsmath}
\usepackage{algorithmic}
\usepackage{xcolor, color}
\usepackage{array, makecell, multirow}
\usepackage{amssymb}
\journal{Pattern Recognition}

\begin{document}

\begin{frontmatter}



\title{Detection and Rectification of Arbitrary Shaped Scene Texts by using Text Keypoints and Links}


\author[address1]{Chuhui Xue}
\author[address1]{Shijian Lu\corref{mycorrespondingauthor}}
\cortext[mycorrespondingauthor]{Corresponding author \\
                                Email Address: xuec0003@e.ntu.edu.sg, shijian.lu@ntu.edu.sg and shoi@salesforce.com}

\author[address2]{Steven Hoi}

\address[address1]{School of Computer Science and Engineering, Nanyang Technological University, Singapore}
\address[address2]{Salesforce, Singapore}

% \begin{highlights}
% \item We propose a robust scene text detection and rectification technique that is capable of detecting and rectifying scene texts of arbitrary shapes almost simultaneously.
% \item We formulate scene text detection and rectification as a text keypoint and link detection problem and proposes a mask-guided multi-task network that is capable of detecting text keypoints and keypoint links accurately.
% \item We develop an efficient and end-to-end trainable system that achieves superior scene text detection and rectification performance as compared with the state-of-the-art.
% \end{highlights}

\begin{abstract}
Detection and recognition of scene texts of arbitrary shapes remain a grand challenge due to the super-rich text shape variation in text line orientations, lengths, curvatures, etc. This paper presents a mask-guided multi-task network that detects and rectifies scene texts of arbitrary shapes reliably. Three types of keypoints are detected which specify the centre line and so the shape of text instances accurately. In addition, four types of keypoint links are detected of which the horizontal links associate the detected keypoints of each text instance and the vertical links predict a pair of landmark points (for each keypoint) along the upper and lower text boundary, respectively. Scene texts can be located and rectified by linking up the associated landmark points (giving localization polygon boxes) and transforming the polygon boxes via thin plate spline, respectively. Extensive experiments over several public datasets show that the use of text keypoints is tolerant to the variation in text orientations, lengths, and curvatures, and it achieves superior scene text detection and rectification performance as compared with state-of-the-art methods.
\end{abstract}


\end{frontmatter}


\section{Introduction}
\label{sec:intro}
Scene text understanding has attracted increasing interest in the community of computer vision as texts in scenes often convey rich semantic information that is important to various applications such as image search, autonomous driving, etc. Quite a number of scene text detection works have been reported recently by using Convolutional Neural Network (CNN) which have achieved very promising result. One typical approach \cite{zhou2017east,Wang_2019_CVPR} treats word/line-level instances as generic objects but it often faces various problems while handling arbitrary-shaped texts that have super-rich variation in text line orientations, lengths, curvatures, etc. Another approach \cite{hu2017wordsup,Baek_2019_CVPR} works in the bottom-up manner by first detecting character candidates and then linking them up for word/text-line detection. On the other hand, this approach neglects global information at word/text-line level which often leads to sub-optimal detection. 

At the other end, perspective and curvature distortions in many arbitrary-shaped scene texts often affect the scene text recognition performance greatly. Scene text rectification which aims to restore scene texts to a straight and fronto-parallel view is critical for accurate recognition of texts of arbitrary shapes in scenes. Quite a few works have been reported \cite{shi2016robust,shi2018aster} which focus on the rectification scene texts with various geometric distortions by detecting certain feature points or lines. Nevertheless, most existing works treat scene text rectification as an independent problem and design dedicated solutions but neglect the connection between scene text detection and scene text rectification.

\begin{figure}
  \centering
  \includegraphics[width=\linewidth]{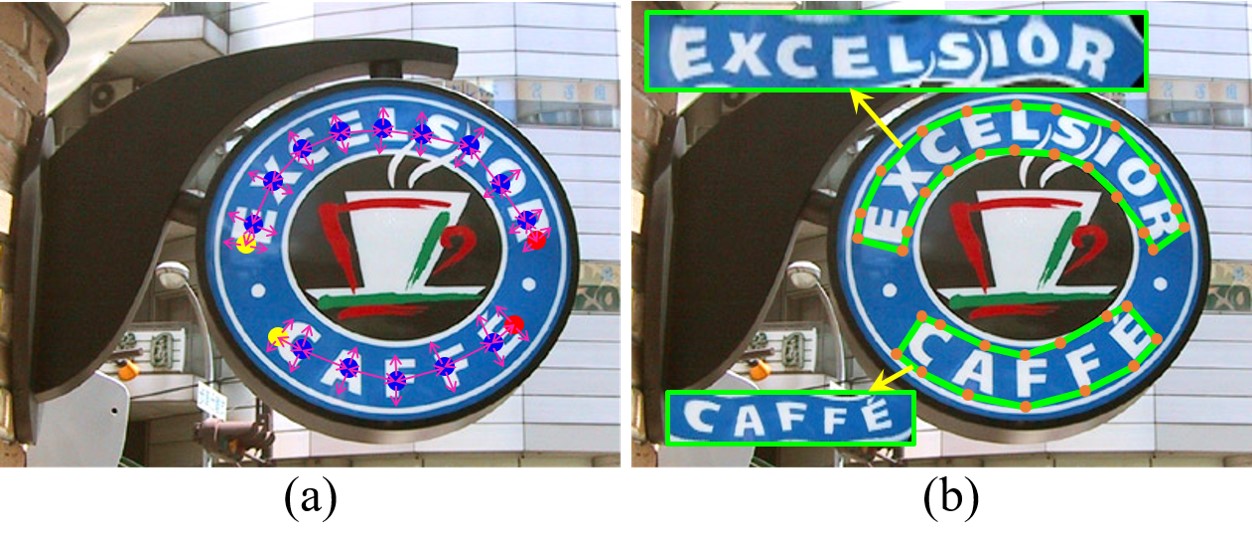}
\caption{Scene text detection and rectification by using text keypoints: The proposed technique detects three types of text keypoints along the text centre line and four types of keypoint links as illustrated in 1a. Of the detected keypoint links, the horizontal links associate text keypoints to relevant text instances, and the vertical links predict a pair of landmark point (for each keypoint) along the upper and lower text boundary. Scene texts can be located and rectified by linking up the associated landmark points (giving detection polygon boxes) and transforming each polygon box via thin plate spline, respectively, as illustrated in 1b.}
\label{fig:overview}
\end{figure}

This paper presents a mask-guided multi-task network that locates and rectifies scene texts by detecting text keypoints and keypoint links. Three types of text keypoints are detected at each character and the two ends of text instances as highlighted by blue, yellow and red points in Fig. \ref{fig:overview}a which define the text line shapes accurately. Unlike existing works that take a bottom-up approach, our proposed technique captures global instance-level information and is capable of deriving ground-truth text keypoints and keypoint links from the original word/line-level annotations with no extra character annotation. In addition, four types of keypoint links are detected at each text keypoints of which the horizontal links associate the detected text keypoints of each text instance and the vertical links predict a pair of landmark points along the upper and lower text boundaries, respectively, as illustrated in Fig. \ref{fig:overview}a. Finally, scene texts can be located and rectified by linking up the associated land-mark points (giving localization polygon boxes) and transforming the polygon boxes via thin plate spline \cite{bookstein1989principal} (TPS), respectively, as illustrated in Fig. \ref{fig:overview}b. Unlike existing scene text detection and rectification works that often treat scene text detection and scene text rectification as separate tasks, our proposed technique detects and rectifies arbitrary-shaped scene text almost simultaneously, where scene text rectification can be achieved with little extra efforts.

The contribution of this work are three-fold. First, it proposes a robust scene text detection and rectification technique that is capable of detecting and rectifying scene texts of arbitrary shapes almost simultaneously. Second, it formulates scene text detection and rectification as a text keypoint and link detection problem and proposes a mask-guided multi-task network that is capable of detecting text keypoints and keypoint links accurately. Third, it develops an efficient and end-to-end trainable system that achieves superior scene text detection and rectification performance as compared with the state-of-the-art.

The rest of this paper is organized as follows. Section \ref{sec:related} gives a brief review of related works on scene text detection and rectification. In Section \ref{sec:method}, we present the proposed scene text detector in details including the problem formulation, scene text detection, scene text rectification and network training. Section \ref{sec:experiment} shows quantitative and qualitative experimental results. Section \ref{sec:conclusion} concludes this paper.

%--------------------------------------------------------
\section{Related Work}
\label{sec:related}

\subsection{Scene Text Detection}
The CNN-based works detect scene texts via three typical approaches. The first approach \cite{sihang2020precise,wang2019efficient,xu2019geometry,Wang_2020_CVPR,zhu2020textmountain,he2020realtime,liu2019curved} treats scene texts as a specific class of objects and follows the pipeline of generic object detection that detects word/line-level texts directly by using Faster RCNN \cite{ren2015faster}, SSD \cite{liu2016ssd}, DenseBox \cite{huang2015densebox}, etc. This approach either designs text-specific proposals/default boxes to fit the shapes of text instances \cite{shi2017detecting,Liu_2017_CVPR,he2017single,liao2018rotation,yang2018inceptext,xie2019scene,Lyu_2018_ECCV,Zhang_2019_CVPR,Wang_2019_CVPR,tian2016detecting,liao2018textboxes++}, or regresses text pixels to the boundaries/vertices of text instances \cite{He_2017_ICCV,zhou2017east,liu2018fots,Long_2018_ECCV,Xue_2018_ECCV,msr_2019_ijcai}. The second approach \cite{Liu_2018_CVPR,Liu_2019_CVPR} treats scene text detection as an instance segmentation problem which first classifies image pixels to text/non-text class and then groups text pixels to generate text bounding boxes by exploiting predicted links \cite{deng2018pixellink}, embedding \cite{Tian_2019_CVPR}, text borders \cite{wu2017self}, progressive expansion \cite{Wang_2019_CVPR}, etc. The third approach \cite{tian2017wetext,hu2017wordsup,Baek_2019_CVPR,tang2019seglink++} takes the bottom-up approach which first detects characters and then groups them into words or lines.

Our proposed technique detects both character-level and instance-level keypoints which is more robust to orientations, lengths and shapes of text instances and leads to more accurate detection of multi-oriented, long and arbitrary-shaped text instances. The landmarks on the boundary of text bounding boxes are hence predicted by the detected keypoints as well as keypoint links, which simply form the main structure of text instances and can be easily applied as control points for text rectification. The overall network is trained in a regular manner where additional training steps are not required.

\subsection{Scene Text Rectification}
Scene text rectification has attracted increasing attention in recent years due to its significance and effectiveness for recognizing texts in arbitrary shapes. Quite a few works explored rectification for document image understanding \cite{jagannathan2005perspective,lu2005perspective,lu2006document}. Recently, \cite{yang2017learning} proposed an auxiliary character detection model and an alignment loss for precise character localization. \cite{bartz2018see,shi2016robust,shi2018aster,liu2018char} incorporate scene text rectification for better scene text recognition. \cite{Zhan_2019_CVPR} presented line-fitting transformation with iterative rectification which achieved promising performance.

The recent works treat scene text rectification as an independent task, whereas the connection between scene text detection and scene text rectification is largely neglected. Our proposed technique detect and rectify scene texts simultaneously by detecting text keypoints and predicting text landmark points that are more tolerant to the variation in text line shapes. The rectified texts are also more ``readable" which improve the scene text recognition performance significantly.

%-------------------------------------------------------------------------
\section{Methodology}\label{sec:method}

\begin{figure}[!t]
  \centering
  \includegraphics[width=\linewidth]{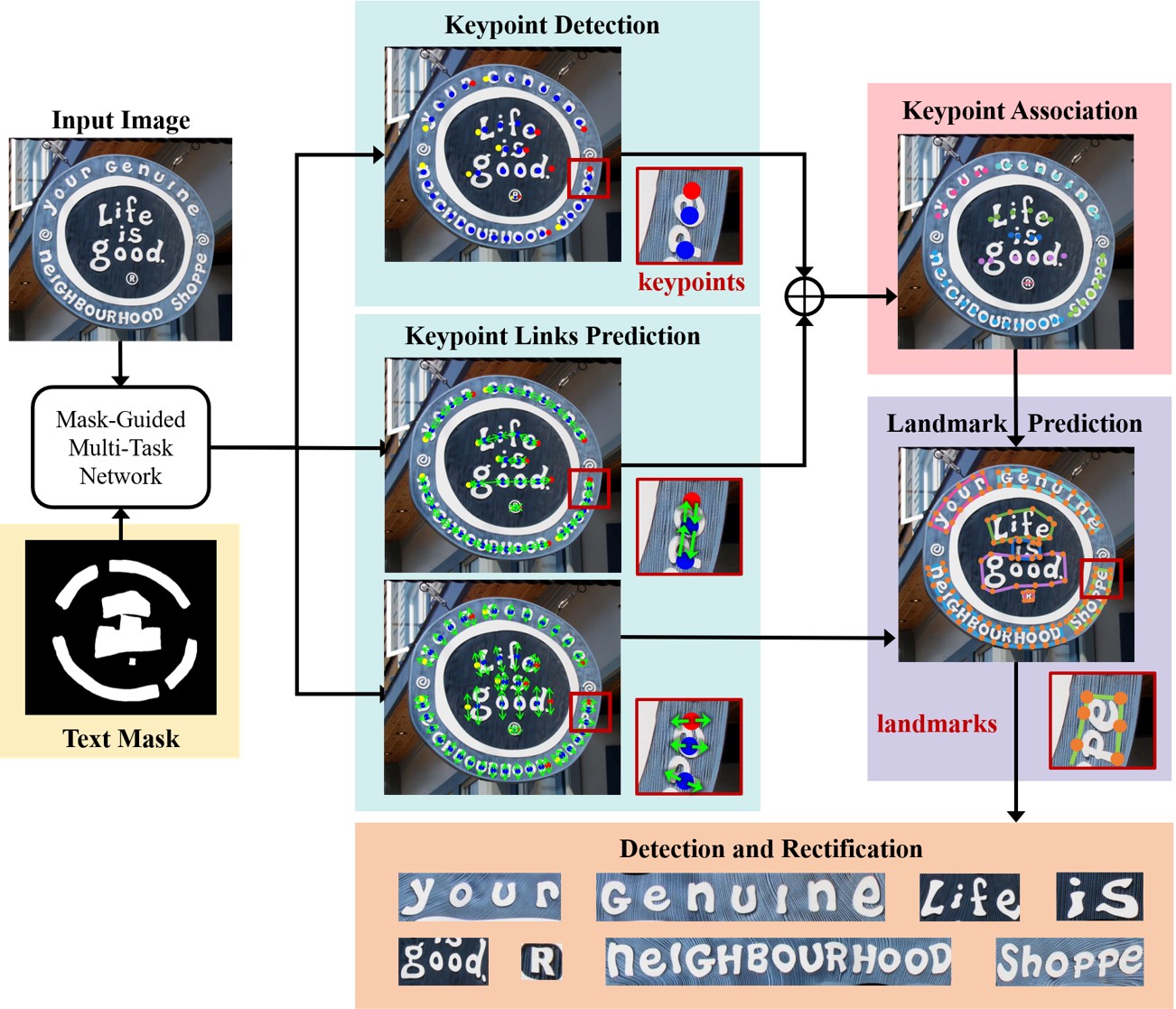}
\caption{The framework of the proposed technique: A mask-guided multi-task network is designed which detects three types of text keypoints (as highlighted in blue, yellow and red points in \textbf{Keypoint Detection}) and four types of keypoint links (as highlighted in green color arrows in \textbf{Keypoint Links Prediction}), respectively. The horizontal links associate text keypoints of the same text instance as illustrated in \textbf{Keypoint Association}, and the vertical links predict a pair of landmark points (for each text keypoint) along the upper and lower text boundary as illustrated in \textbf{Landmark Prediction}. Finally, scene texts can be detected and rectified by linking up the associated landmark points (giving text localization polygon boxes) and transforming each polygon box via simple template transformation as illustrated \textbf{Detection and Rectification}, respectively. }
\label{fig:framework}
\end{figure}

\subsection{Problem Formulation}
\subsubsection{Keypoints} 
\begin{figure}[t]
  \centering
  \includegraphics[width=\linewidth]{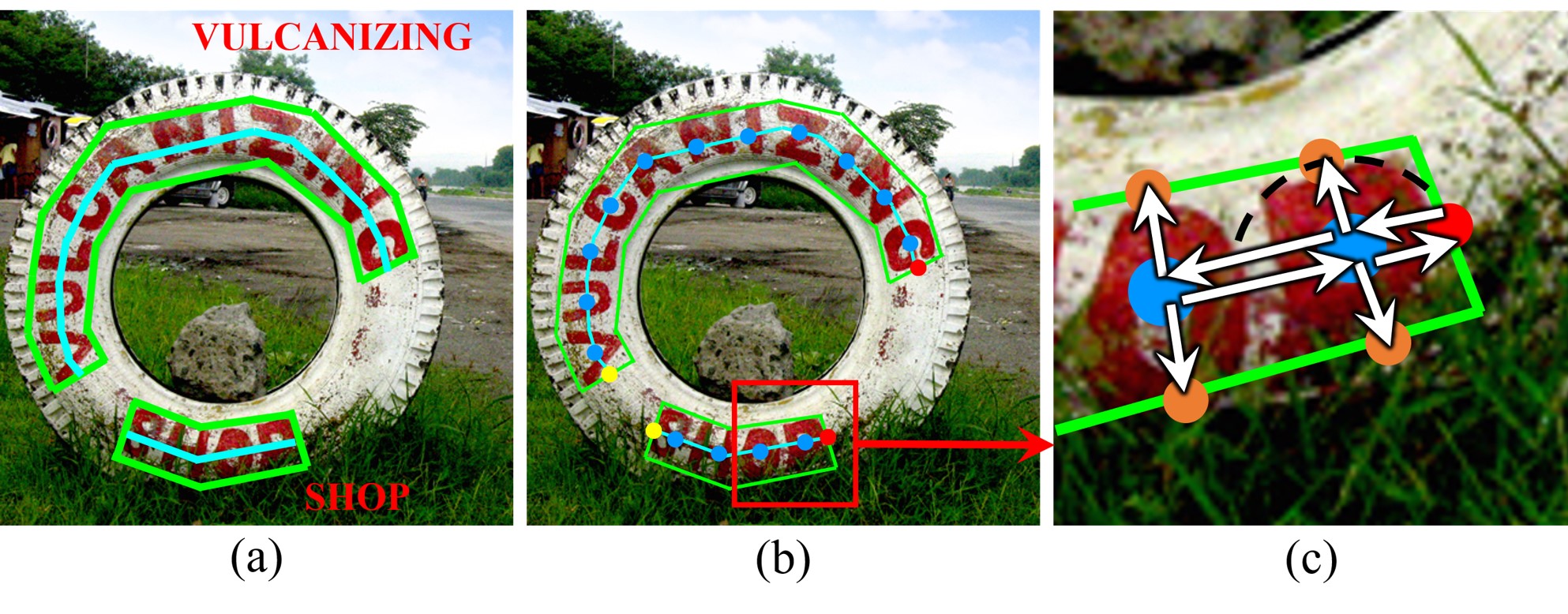}
\caption{The generation of ground-truth text keypoints and links: Given text annotation polygons and the transcription as shown in 3a, the text centre line and the text keypoints at characters and left and right text boundary can be derived as highlighted in blue, yellow and red points in 3b. For each text keypoint, a pair of text landmark points can be derived as the nearest points at the upper and lower text boundaries as highlighted by orange points in 3c, and the four types of keypoint links can be derived by line segments that start from the current text keypoint and end at the neighbouring text keypoints on the left and right as well as the associated landmark points at the upper and lower text boundary as highlighted by white arrows in 3c, respectively.}
\label{fig:kp_gen}
\end{figure}

We formulate the scene text detection as a text keypoint detection and association problem. Three types of text keypoints are defined which include: 1) character keypoint that is located along the text middle line around characters in each text instance, 2) left keypoint that is located at the left end boundary of each text instance and 3) right keypoint that is located at the right end boundary of each text instance. The blue, yellow and red points in Fig. \ref{fig:kp_gen} illustrate the three types of keypoints. The character keypoints are designed to capture local features at character or stroke level, where the left and right keypoints also provide boundary information of each text instance which is very useful in the keypoint association.

During the network training stage, the three types of keypoints can be directly generated from the original image annotations including the bounding box and the transcription of each text instance. Given the annotation of a text instance, the cyan-color centre line and the number of characters $n$ can be simply derived from the original green-color text bounding boxes and the text transcription, respectively, as illustrated in Fig. \ref{fig:kp_gen}. The $n$ blue-color character keypoints can thus be generated so that they all lie on the centre line of the text instance and have the same distance between neighbouring character keypoints. The left and right keypoints can be located at the centre of two vertical boundaries at the left and right ends of the text instance. The yellow and red points in Fig. \ref{fig:kp_gen}b illustrate the left and right keypoints, respectively. Note the character keypoints may not lie exactly at the centre of each character due to the different widths of different characters, and our detection network is not a character detector which does not require accurate character-level annotations.

\subsubsection{Landmark Points} A set of landmark points are defined for each text instance for the generation of text bounding box and rectification of distorted texts. For each text keypoint, a pair of landmark points are located at the upper and lower text boundaries that have the closest distance to the text keypoint of interest as highlighted by orange-color points in Fig. \ref{fig:kp_gen}c. The landmark points once detected can be exploited to not only produce the text bounding boxes but also rectify distorted texts via thin plate transformation. As the landmark points are determined by text keypoints lying around each character, scene text rectification by using landmark points is less sensitive to the scene text variation in styles, orientations, shapes, etc. 

\subsubsection{Keypoint Links} Four types of text keypoint links are defined for associating keypoints within text instances and predicting the landmark points for scene text detection and rectification. Of the four types of keypoint links, two types are vertical which are defined by the distance between each text keypoint and its associated landmark points at the upper and lower text boundary as highlighted by white-color arrows in Fig. \ref{fig:kp_gen}c. The other two types are horizontal which are defined by the distance between each text keypoint and its neighbouring text keypoints one the left and right as illustrated by white-color arrows in Fig. \ref{fig:kp_gen}c.

\begin{figure}[th!]
  \centering
  \includegraphics[width=\linewidth]{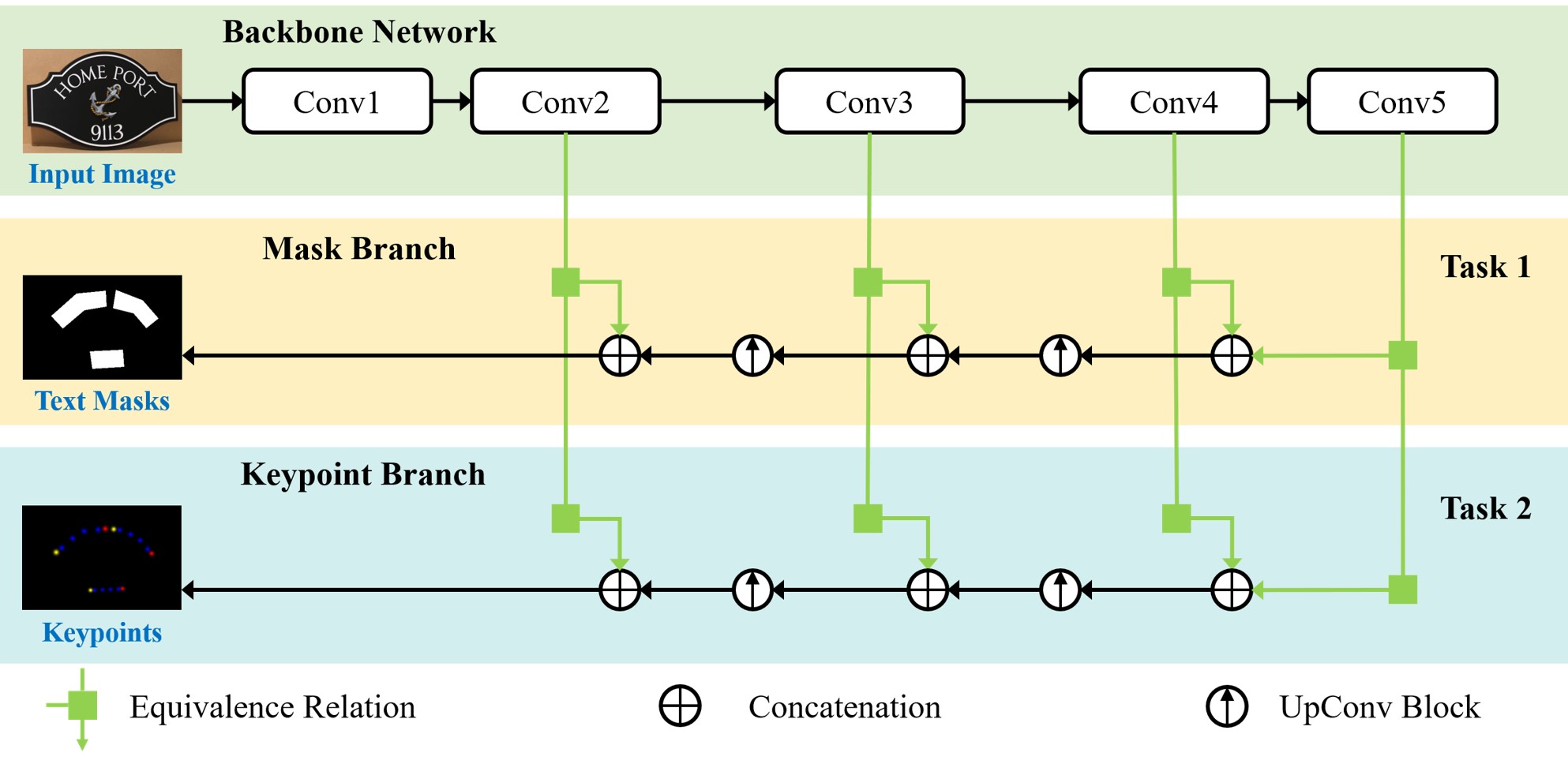}
\caption{Architecture of the proposed mask-guided multi-task detection network: Two feature fusion branches are designed and incorporated for the mask segmentation and keypoint detection respectively. The two branches fuse the features from layers \textit{Conv2} - \textit{Conv5} of the U-Net backbone network \cite{ronneberger2015u}, where the mask branch serves as a global guidance which guides the network to learn shape features at text instance level whereas the keypoint branch learns local character features for the detection of text keypoints and keypoint links.}
\label{fig:network}
\end{figure}

\subsection{Scene Text Detection}
\subsubsection{Keypoint and Keypoint Link Detection }
Accurate and robust detection of text keypoints and keypoint links is critical for the ensuing scene text detection and rectification tasks. We design a mask-guided multi-task network which fuses local and global features for accurate detection of text keypoints and keypoint links as illustrated in Fig. \ref{fig:network}. As Fig. \ref{fig:network} shows, two feature fusion branches are designed to fuse the features from layers \textit{Conv Stage2} - \textit{Conv Stage 5} of the U-Net \cite{ronneberger2015u} architecture. Specifically, the \textbf{Mask Branch} focuses on the text mask segmentation and the \textbf{Keypoint Branch} focuses on keypoint and keypoint link detection, respectively.

In the \textbf{Mask Branch}, a text region segmentation map is produced where the pixels within the text bounding boxes will have positive responses. The text segmentation map just serves to guide the network to learn global information at text instance level, instead of just focusing on local features at character levels. It will not be passed to the following inference steps.

In the \textbf{Keypoint Branch}, three heat-maps are predicted for the detection of three types of keypoints around characters and left and right text instance boundary, respectively. As text keypoints generated from instance-level annotations may not lie exactly at the centre of each character, we reduce penalty on negative locations within a radius of positive locations (instead of applying equal penalty as CornerNet \cite{law2018cornernet}). Specifically, the penalty radius for each text keypoint is determined by the distances $d$ between the associated landmark points at the upper and lower text boundaries. With the distance $d$, the penalty reduction is determined by an normalized 2D Gaussian that has its centre at the text keypoint location and a $\sigma$ of $\sqrt{d}$. 

In addition, the \textbf{Keypoint Branch} predicts eight heat-maps for the four types of keypoint links which contain the horizontal and vertical distances (in x and y coordinates) between each text keypoint and its two neighbouring text keypoints on the left and right as well as its two associated landmark points at the upper and lower text boundaries, respectively. 

\subsubsection{Keypoint Association} \label{sec:kp_assoc}
For each scene text image, the proposed mask-guided multi-task network will detect a set of text keypoints as well as four keypoints links (three links for keypoints at the end of left/right text boundaries) for each text keypoint that points to the two neighbouring text keypoints and two landmark points at the upper and lower text boundaries, respectively. The detected text keypoints can thus be associated and grouped to text instances by using a directed graph $G$. In the graph $G$, any two keypoints will be connected and grouped together if they have a keypoint link between them. In addition, the text keypoints at the left/right text boundary tell the start and end of a text instance. 

\begin{figure}[!t]
  \centering
  \includegraphics[width=\linewidth]{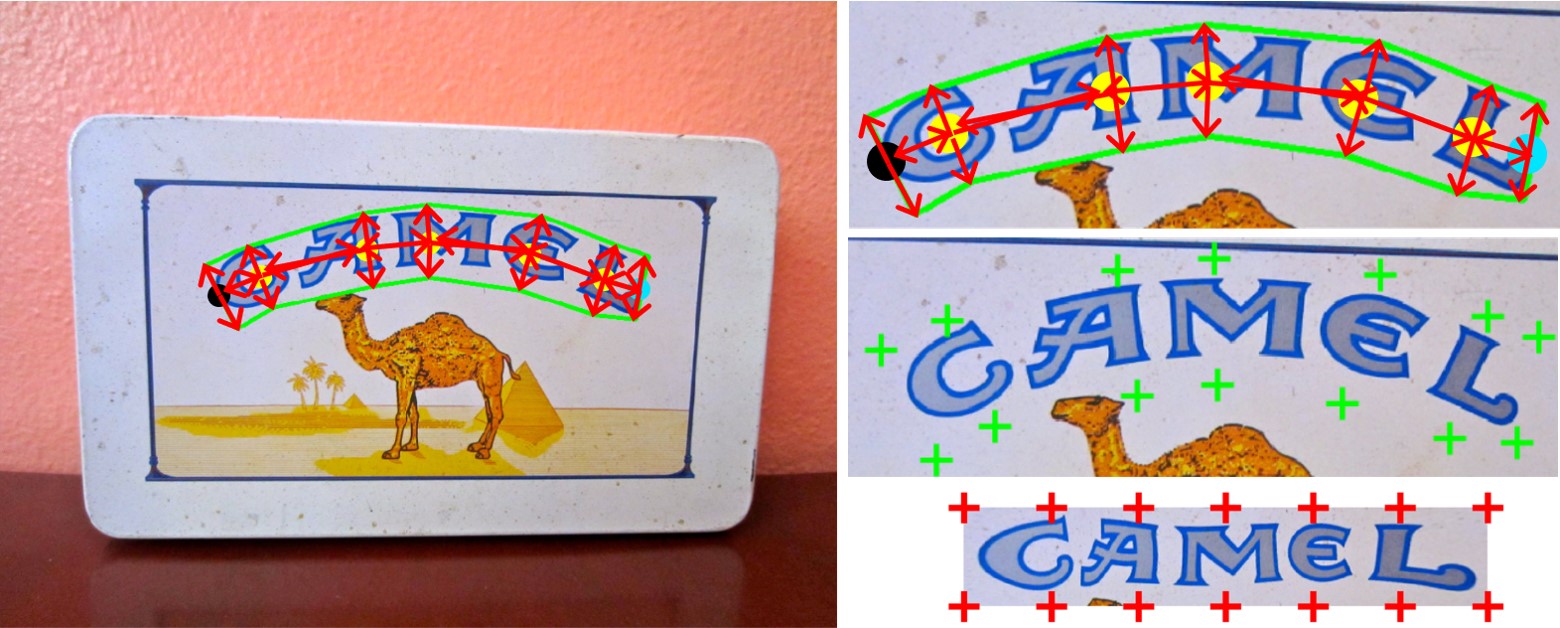}
\caption{Illustration of the text rectification: With the detected text Keypoints (black, yellow and cyan dots) and Keypoint Links (red arrows), the Landmark Points (green crosses) of text instances are obtained. The text instances are rectified by TPS transformation with control points from source (green crosses) to target (red crosses).
}
\label{fig:tps}
\end{figure}

Specifically, given a scene text image, our network detects a set of keypoints each of which is associated with four links that point to upwards, downwards, leftwards and rightwards, respectively. Only the two links pointing to leftwards and rightwards are used in keypoint grouping. Take a text images with a word ‘CAMEL’ as an example as shown in Fig. \ref{fig:tps}. Our network will detect five character keypoints $p_C$, $p_A$, $p_M$, $p_E$ and $p_L$ around the center of the five letters. For one character such as `A', it will have two horizontal links $v_A^l$ and $v_A^r$ that point from $p_A$ to two points $p_A^l$ and $p_A^r$ (two end points of $v_A^l$ and $v_A^r$ that lie on the left and right of $p_A$), respectively. Ideally, $p_A^l$ and $p_A^r$ should overlaps with $p_C$ and $p_M$ perfectly but that seldom happens due to prediction errors.

The neighboring letter that lies on the right (or left) of the letter `A' can thus be identified based on the distances from $p_A$ and $p_A^r$ (or $p_A^l$) to the keypoint of the rest letters (i.e. $p_C$, $p_M$, $p_E$ and $p_L$ in this illustration sample). Take letter `M' as an example. We define a distance ratio $\frac{D_{p_A^r \rightarrow p_M}}{D_{p_A \rightarrow p_M}}$ (or $\frac{D_{p_A^l \rightarrow p_M}}{D_{p_A \rightarrow p_M}}$), where $D_{p_A^r \rightarrow p_M}$ (or $D_{p_A^l \rightarrow p_M}$) and $D_{p_A \rightarrow p_M}$ denote the distances from $p_A^r$ ($p_A^l$) and $p_A$ to $p_M$, respectively. For the correct rightward link to letter `M', $D_{p_A^r \rightarrow p_M}$ is usually very small (as $p_A^r$ is close to $p_M$) and so does the distance ratio. But for false rightward links, e.g. to letter `L’, $D_{p_A^r \rightarrow p_L}$ is large and so does the distance ratio. The correct rightward linked letter can thus be identified by using a distance ratio threshold (set at 0.5 in our system). The leftwards-linked letter can be identified in the similar way.

\subsection{Scene Text Rectification}
Scene text images can be simply rectified based on the associated text keypoints as described in Sec. \ref{sec:kp_assoc}. For each text keypoint with a determined subgraph $G$, its two associated landmark points at the upper and lower text boundaries can be determined by using the two predicted vertical keypoint links. The text keypoints (and so the associated landmark points) with $G$ can then be sorted according to the predicted horizontal keypoint links by topological sorting. Finally, the distorted scene text instance can be rectified by TPS transformation where the detected landmark points can serve a transformation control points.

In implementation, the predicted landmark points are first extended upwards and downwards a bit so as to include the whole text region. The extension is along the direction of the predicted vertical keypoint links with a certain amount that is proportional to the length of the vertical keypoint links (0.1 in our system) as illustrated by the green crosses in the top right image in Fig. \ref{fig:tps}. The mapping correspondence can thus be constructed between the expanded landmark points and a set of target control points which lie on the upper and lower edges of a horizontal rectangle bounding box that has the same height and width as the detected text bounding box as illustrated by the green crosses in the bottom right image in Fig. \ref{fig:tps}. With the established control point correspondences, distorted scene texts can be finally rectified by applying TPS transformation as illustrated in Fig. \ref{fig:tps}.

\subsection{Network Training}
The training of the proposed mask-guided multi-task network requires a list of inputs including the original image, the corresponding binary text region mask, text keypoints, keypoint links, and refined text keypoints. The training aims to minimize the following multi-task loss functions:

\begin{equation}
\mathcal{L}=  \mathcal{L}_{kp} + \lambda_{mask} * \mathcal{L}_{mask} +  \lambda_{link} * \mathcal{L}_{link}
\end{equation}
\noindent where $\mathcal{L}_{kp}$, $\mathcal{L}_{mask}$ and $\mathcal{L}_{link}$ denote the loss of text keypoint detection, mask segmentation, keypoint link prediction and local refinement module, respectively. Parameters $\lambda_{mask}$ and $\lambda_{link}$ are the weight to balance the four losses which are empirically set at 0.1 and 1.0 in our system.

For the loss of text keypoints $\mathcal{L}_{kp}$, we adopt the variant of Focal Loss \cite{lin2017focal,law2018cornernet}:
\begin{equation}
\mathcal{L}_{kp} = \frac{-1}{N} \sum_{i=1}^{H} \sum_{j=1}^{W} 
    \begin{cases}
      (1 - p_{ij})^{\alpha} log(p_{ij}), & \text{if}  y_{ij} = 1.\\
      (1 - y_{ij})^{\beta} p_{ij}^{\alpha} log(1 - p_{ij}), & \text{otherwise}.
    \end{cases}
\end{equation}
\noindent where $N$ is the number of keypoints, $H$ and $W$ are the height and width of input images, $p_{ij}$ and $y_{ij}$ are the scores at localtion (i, j) of the predicted and the groundtruth heatmaps, and $\alpha$ and $\beta$ are two hyper-parameters which are empirically set to 2 and 4, respectively.

In addition, we apply standard cross-entropy loss \cite{xie2015holistically} for mask prediction task $\mathcal{L}_{mask}$ and standard Smooth L1 loss \cite{girshick2015fast} for link prediction $\mathcal{L}_{link}$.

%------------------------------------------------------------------------
\section{Experiments}\label{sec:experiment}
The proposed technique has been extensively evaluated over four public datasets that contain scene texts of arbitrary orientations, lengths, and shapes. It has also been compared with state-of-the-art techniques and analyzed via ablation analysis to be described in the ensuing subsections.

\begin{table}[!t]
\centering
 \caption{Comparison of our proposed scene text detection method with state-of-the-art methods over dataset \textbf{Total-Text}: For fair comparisons, all compared methods do not use extra images in training and use single-scale evaluations in testing.}
 \begin{tabular}{|c | c | c | c | c|} 
 \hline
 \textbf{Methods} & \textbf{Recall} & \textbf{Precision} & \textbf{F-score} & \textbf{FPS} \\ 
 \hline\hline
 PSENet \cite{psenet}                       & 75.1              & 81.8              & 78.3          & 3.9 \\
 TextSnake \cite{Long_2018_ECCV}            & 74.5              & 82.7              & 78.4          & - \\
 Wang \textit{et al.}\cite{Wang_2019_CVPR}  & 76.8              & 80.9              & 78.9          & - \\
 MSR \cite{msr_2019_ijcai}                  & 74.8              & 83.8              & 79.0          & 1.1 \\
 CSE \cite{Liu_2019_CVPR}                   & 79.1              & 81.4              & 80.2          & 2.4 \\
 TextField \cite{xu2019textfield}           & 79.9              & 81.2              & 80.6          & - \\
 LOMO \cite{Zhang_2019_CVPR}                & 75.7              & \textbf{88.6}     & 81.6          & -\\
 \hline\hline
 \textbf{Ours}                              & \textbf{82.6}     & 86.1             & \textbf{84.4} & \textbf{7.2} \\
 \hline
 \end{tabular}
 \label{tab:total}
\end{table}

\begin{table}[!t]
 \caption{Comparison of our proposed scene text detection method with state-of-the-art methods over dataset \textbf{CTW1500}: For fair comparisons, all compared methods do not use extra images in training and use single-scale evaluations in testing.} 
\centering
 \begin{tabular}{|c | c | c | c | c|} 
 \hline
 \textbf{Methods} & \textbf{Recall} & \textbf{Precision} & \textbf{F-score} & \textbf{FPS} \\  
 \hline\hline
 TextSnake \cite{Long_2018_ECCV}            & 67.9              & 85.3              & 75.6          & - \\
 PSENet \cite{psenet}                       & 75.6              & 80.0              & 78.0          & 3.9 \\
 LOMO \cite{Zhang_2019_CVPR}                & 69.6              & \textbf{89.2}     & 78.4          & - \\
 CSE \cite{Liu_2019_CVPR}                   & 76.0              & 81.1              & 78.4          & 2.6 \\
 Embedding \cite{Tian_2019_CVPR}            & 77.8              & 82.7              & 80.1          & - \\
 Wang \textit{et al.}\cite{Wang_2019_CVPR}  & \textbf{80.2}     & 80.1              & 80.1          & - \\
 TextField \cite{xu2019textfield}           & 79.8              & 83.0              & 81.4          & - \\
 MSR \cite{msr_2019_ijcai}                  & 78.3              & 85.0              & 81.5          & 1.1\\
 \hline\hline
 \textbf{Ours}                              & 77.7              & 88.3              & \textbf{82.7} & \textbf{7.6} \\
%  \hline\hline
%  MSR* \cite{msr_2019_ijcai}                 & 79.3              & 84.5              & 81.8          & 1.1\\
%  PSENet* \cite{psenet}                      & 78.7              & 85.6              & 82.0          & 5.4 \\
%  \textbf{Ours*}                             & \textbf{82.8}     & \textbf{85.9}     & \textbf{84.3} & \textbf{7.6} \\
 \hline
 \end{tabular}
 \label{tab:ctw}
\end{table}

\subsection{Datasets}
\noindent \textbf{SynthText} \cite{Gupta16} contains more than 800,000 synthetic scene text images most of which are at both word and character level with multi-oriented rectangular annotations.

\noindent \textbf{CTW1500} \cite{yuliang2017detecting} consists of 1,000 training images and 500 test images that contain 10,751 multi-oriented text instances of which 3,530 are arbitrarily curved. Most of the text instances are annotated at text-line level by using 14 vertices, where texts are largely in English and Chinese.

\noindent \textbf{Total-Text} \cite{ch2017total} consists of 1,255 training images and 300 test images where texts are all in English. It contains a large number of multi-oriented curved text instances each of which is annotated at word level by using a polygon. 

\noindent \textbf{ICDAR2019-ArT} \cite{chng2019icdar2019} is a combination of CTW1500, Total-Text and Baidu Curved Scene Text datasets which consists of 5,603 training images and 4,563 test images. It contains a large amount of texts in arbitrary shapes which are annotated at word level by polygons. 

\noindent \textbf{MSRA-TD500} \cite{yao2012detecting} consists of 300 training images and 200 test images. All captured text instances are printed in English and Chinese which are annotated at text-line level by using best-aligned rectangles. 

\subsection{Evaluation Metrics}
The evaluation of the proposed method follows recent scene text detection techniques \cite{msr_2019_ijcai,psenet,Long_2018_ECCV} and scene text datasets \cite{yuliang2017detecting,ch2017total,chng2019icdar2019}. Specifically, IoU-based evaluation protocol is adopted for evaluation of the proposed technique. We first calculate the Intersection over Union (IoU) of detected text instances with the corresponding ground-truth. The text instances with IoU larger than 0.5 is regarded as match and counted as True Positive (TP). In case of multiple matches, only the detection region with the highest IoU is counted as TP while the rest matches will be counted as False Positive (FP). Other detection are counted as False Negative (FN). Precision, Recall, and F-score (F) are calculated by:

\begin{equation}
Precision = \frac{TP}{TP+FP}
\end{equation}

\begin{equation}
Recall = \frac{TP}{TP+FN}
\end{equation}

\begin{equation}
F = \frac{2 \times Precision \times Recall}{Precision+Recall}
\end{equation}

\subsection{Implementation Details}
The proposed technique is implemented using Tensorflow \cite{abadi2016tensorflow} on a regular GPU workstation with 2 Nvidia Geforce GTX 1080 Ti, an Intel(R) Core(TM) i7-7700K CPU @ 4.20GHz and 32GB RAM. The network is optimized by Adam optimizer \cite{kingma2014adam} with a starting learning rate of $10^{-4}$. The network is pre-trained on the SynthText \cite{Gupta16}, which is then fine-tuned by using the training images of each evaluated dataset with a batch size of 10. ResNet-50 \cite{he2016deep} is used as the network backbone.

\subsection{Experimental Results}
The proposed technique has been evaluated quantitatively and qualitatively for scene text detection, rectification and recognition tasks over four public datasets as shown in Tables 1-5 and Figs. 5-7. It has also been analyzed through ablation studies as shown in Table \ref{tab:ablation}.

\begin{table}[!t]
 \caption{Comparison of our proposed scene text detection method with state-of-the-art methods over dataset \textbf{ICDAR2019-Art}: For fair comparisons, all compared methods do not use extra images in training and all use single-scale evaluations in testing.}
\centering
 \begin{tabular}{|c | c | c | c | c |} 
 \hline
 \textbf{Methods} & \textbf{Recall} & \textbf{Precision} & \textbf{F-score} & \textbf{FPS}\\ 
 \hline\hline
 PSENet \cite{psenet}                       & 54.1              & \textbf{80.1}     & 64.4              & 5.4\\
 MSR \cite{msr_2019_ijcai}                  & 56.4              & 75.4              & 64.6              & 1.1\\
 \hline\hline
 \textbf{Ours}                              & \textbf{59.6}     & 77.8              & \textbf{67.5}     & \textbf{7.6}\\
 \hline
 \end{tabular}
 \label{tab:art}
\end{table}

\subsubsection{Scene Text Detection}

\noindent \textbf{Total-Text:} The proposed technique has been evaluated over the dataset Total-Text and Table \ref{tab:total} shows experimental results. For a fair comparison, we only list the recent methods that do not use real images in pre-training and use a single-scale image in testing. As Table \ref{tab:total} shows, the proposed technique achieves a f-score of 84.4\% which outperforms the best f-score by 2.8\%, demonstrating the superiority of the proposed technique on detecting arbitrary-shaped scene texts with word-level annotations.

\begin{table}[!t]
 \caption{Comparison of our proposed scene text detection method with state-of-the-art methods over the dataset \textbf{MSRA-TD500}}
\centering
 \begin{tabular}{|c | c | c | c|} 
 \hline
 \textbf{Methods}                           & \textbf{Recall}   & \textbf{Precision}    & \textbf{F-score} \\  
 \hline\hline
 PixelLink \cite{deng2018pixellink}         & 73.2              & 83.0                  & 77.8 \\
 TextSnake \cite{Long_2018_ECCV}            & 73.9              & 83.2                  & 78.3 \\
 RRD \cite{liao2018rotation}                & 73.0              & 87.0                  & 79.0 \\
 Xue \textit{et al.} \cite{Xue_2018_ECCV}   & 77.4              & 83.0                  & 80.1 \\
 ITN-ResNet50 \cite{Wang_2018_CVPR}         & 72.3              & 90.3                  & 80.3 \\
 TextField \cite{xu2019textfield}           & 75.9              & 87.4                  & 81.3          \\
 Lyu \textit{et al.} \cite{Lyu_2018_CVPR}   & 76.2              & 87.6                  & 81.5 \\ 
 MSR \cite{msr_2019_ijcai}                  & 76.7              & 87.4                  & 81.7          \\
 Embedding \cite{Tian_2019_CVPR}            & 81.7              & 84.2                  & 82.9           \\
 MCN \cite{Liu_2018_CVPR}                   & 79.0              & 88.0                  & 83.0  \\
 IncepText \cite{yang2018inceptext}         & 79.0              & 87.5                  & 83.0 \\
 Wang \textit{et al.}\cite{Wang_2019_CVPR}  & \textbf{82.1}     & 85.2                  & 83.6          \\
 \hline\hline
 \textbf{Ours}                              & 76.7              & \textbf{92.1}         & \textbf{83.7} \\
 \hline
 \end{tabular}
 \label{tab:msra}
\end{table}

\begin{figure}[!t]
  \centering
  \includegraphics[width=\linewidth]{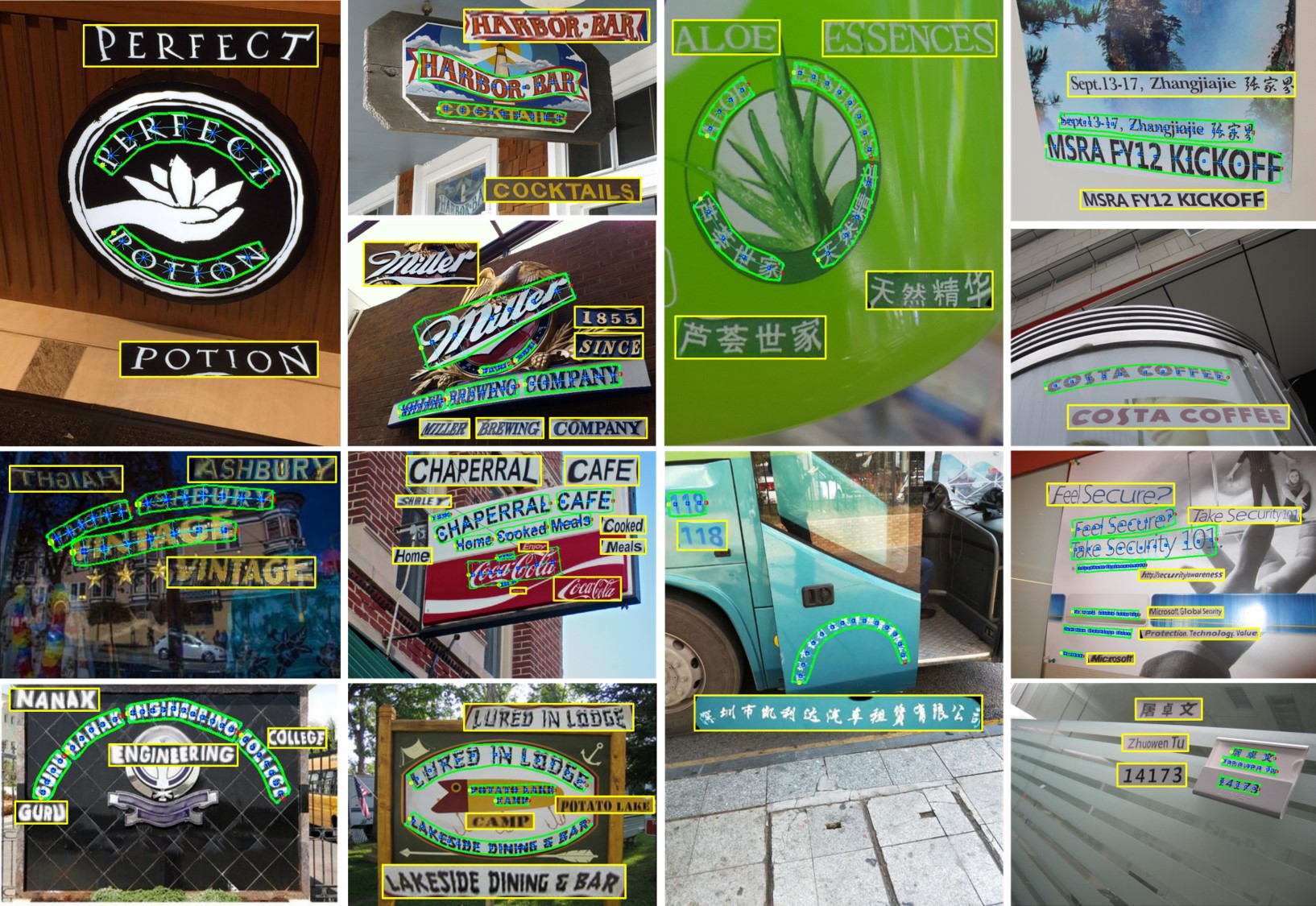}
\caption{Illustration of the proposed scene text detection and rectification method: Sample images in columns 1-4 are selected from Total-Text, CTW1500, ArT and MSRA-TD500, where green boxes highlight the detection polygon boxes, yellow boxes show the rectified text patches by our proposed method. The points in blue, yellow and red highlight the detected text keypoints and the arrows in light blue highlight the detected keypoint links. Zoom-in may be needed for clear view.}
\label{fig:sample}
\end{figure}

\noindent \textbf{CTW1500:} For dataset CTW1500, we get its transcriptions from dataset ICDAR2019-ArT since CTW1500 does not provide the ground-truth transcription. As the annotations in ArT are at word level while those of CTW1500 are largely at text-line level, we generate the text-line level transcriptions for CTW1500 by merging the word level transcriptions in ArT for words that are in the same annotated text line in CTW1500. As Table \ref{tab:ctw} shows, the proposed technique achieves a f-score of 82.7\% which outperforms all state-of-the-art methods. 

\noindent \textbf{ICDAR2019-Art:} We also evaluate the proposed technique over one recent large-scale dataset ICDAR2019-Art. We trained two additional models by using recent scene text detectors including MSR and PSENet for comparisons. All three models are pre-trained by using SynthText without real images and they all use ResNet-50 as backbone and are tested under single-scale manner. As Table \ref{tab:art} shows, the proposed technique achieves state-of-the-art performance that outperforms MSR by 2.9\% in f-score.

\noindent \textbf{MSRA-TD500:} The proposed method has also been evaluated over MSRA-TD500 where most scene texts are straight but in arbitrary orientations without transcription annotations. The purpose is to study how the proposed method performs for straight text lines without ground-truth transcriptions. For each MSRA-TD500 training image, we generate pseudo keypoints by using the model trained on CTW1500 and then use the pseudo keypoints as ground-truth to train the MSRA-TD500 model. Following the state-of-the-art \cite{Lyu_2018_CVPR,Wang_2018_CVPR,Xue_2018_ECCV,Long_2018_ECCV} on this dataset, we include HUST-TR400 \cite{yao2014unified} training images in training. Table \ref{tab:msra} shows experimental results. As Table \ref{tab:msra} shows, the proposed method achieves state-of-the-art performance even though a certain amount of pseudo keypoints are inaccurate or missing.

Fig. \ref{fig:sample} shows a few sample images from datasets Total-Text, CTW1500, ArT and MSRA-TD500, and the scene text detection by our proposed methods. As Fig. \ref{fig:sample} shows, our proposed method is capable of detecting scene texts in various shapes, lengths, and orientation.

\subsubsection{Scene Text Rectification}
We evaluate our proposed scene text rectification technique by running scene text recognition models over the rectified scene text images. The scene text recognition models are trained by using the text image patches that are cropped from the training images of the dataset Total-Text and ArT as shown in Table \ref{tab:rect_comp}. Specifically, we train three ASTER scene text recognition models \cite{shi2018aster} including 1) `ASTER\_No\_Rect' (Baseline) that is trained by using the Total-Text/ArT patches without any rectification; 2) `ASTER\_STN' that is trained by using the Total-Text/ArT patches that are restored by using spatial transformation networks (STN) as proposed in \cite{shi2018aster}; 3) `ASTER\_Ours' that is trained by using the Total-Text/ArT patches that are restored by our proposed rectification method. In testing, all text instances are detected by using our proposed text detector. Similar to the training, `ASTER\_No\_Rect' is applied to the detected texts without rectification, `ASTER\_STN' is applied to STN-restored texts, and `ASTER\_Ours' is applied to the texts rectified by our method. 

\begin{table}[!t]
 \caption{Quantitative comparisons of scene text recognition methods over the dataset Total-Text and ICDAR2019-ArT.}
\centering
 \begin{tabular}{|c | c | c|} 
 \hline
 \multirow{2}{*}{\textbf{Methods}}                        & \textbf{Total-Text}   & \textbf{ArT} \\ \cline{2-3} 
                                                    &  None                 & None(Latin Only)  \\ 
 \hline\hline
 ASTER\_No\_Rect\cite{shi2018aster}                   & 42.0                  & 33.4          \\
 ASTER\_STN\cite{shi2018aster}                    & 46.8                  & 37.4          \\
 \textbf{ASTER\_Ours}                             & \textbf{53.0}         & \textbf{39.9}  \\
 \hline
 \end{tabular}
 \label{tab:rect_comp}
\end{table}

\begin{figure}[t]
  \centering
  \includegraphics[width=\linewidth]{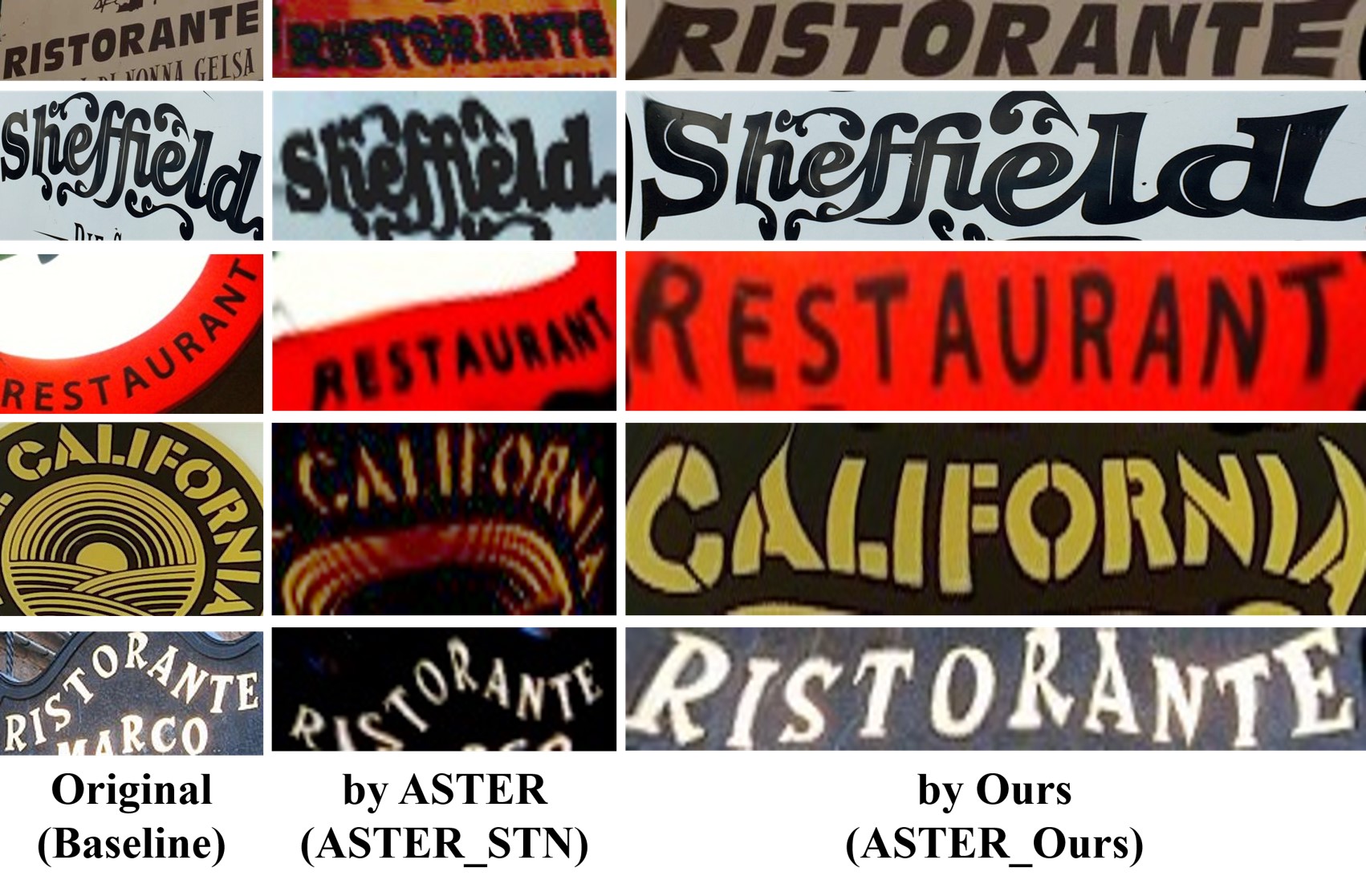}
\caption{Qualitative comparison of scene text rectification methods: For the five scene text images in column 1 that suffer from perspective and curvature distortions, columns 2 and 3 show the rectified images by using ASTER and our proposed method, respectively. Our method produces better rectification clearly.}
\label{fig:rect_comp} 
\end{figure}

Table \ref{tab:rect_comp} shows experimental results. As Table \ref{tab:rect_comp} shows, our rectification helps to improve the recognition performance significantly beyond the Baseline, and it also outperforms the rectification method in ASTER clearly for both datasets. Further, Fig. \ref{fig:rect_comp} shows several sample images from Total-Text. As Fig \ref{fig:rect_comp} shows, our method rectifies scene text images with either perspective (rows 1-2) or curvature (rows 3-5) distortions successfully. Additionally, it performs clearly better than ASTER while handling texts in rare fonts and styles (row 2), with complex background (row 3-4) or in arbitrary shapes (row 5).

\subsubsection{Ablation Study}
As Table \ref{tab:ablation} shows, the baseline `U-Net’ can achieve state-of-the-art detection performance on Total-Text while incorporating our proposed text keypoints and keypoint links. The model `MG U-Net’ that includes the mask-guidance branch improves the detection by 0.7\% in f-score. Further, the model `MGMT Net’ that further includes the multi-task branch (beyond the `MG U-Net’) leads to another 2.1\% improvement in f-score. At the other end, `MGMT Net’ achieves similar FPS as the U-Net although it has a more complex architecture. This shows that the proposed method does not introduce much additional computational cost. % why? - Shijian

\begin{table}[!t]
 \caption{Ablation study of the proposed scene text detection technique over dataset Total-Text: U-Net refers to the original U-Net that detects text keypoints and keypoint links only; MG U-Net refers to the model that incorporates the mask guidance branch beyond KP; MGMT Net refers to the model that incorporate both mask guidance branch and keypoint detection branch.}
\centering
 \begin{tabular}{|c | c | c | c | c|} 
 \hline
 \textbf{Methods} & \textbf{Recall} & \textbf{Precision} & \textbf{F-score} & \textbf{FPS} \\  
 \hline\hline
 U-Net                  & 78.4              & 85.1              & 81.6          & \textbf{8.7} \\
 MG U-Net               & 79.7              & 85.0              & 82.3          & 8.6 \\
 MGMT Net               & \textbf{82.6}     & \textbf{86.1}     & \textbf{84.4} & 7.2 \\
 \hline
 \end{tabular}
 \label{tab:ablation}
\end{table}

As Table \ref{tab:ablation} shows, the proposed scene text detector based on keypoint detection technique can achieve state-of-the-art performance on Total-Text dataset by using U-Net. The inclusion of mask-guidance alone improves the performance of keypoint scene text detector by 0.7\% in f-score. Further, the proposed keypoint detection by the proposed mask-guided multi-task network leads to the best detection performance. On the other hand, although the proposed mask-guided multi-task network is more complex as compared with original backbone network, the processing speed of the proposed technique is only slightly slower than the original U-Net, demonstrating that the proposed technique does not introduce too much additional computational cost.
\begin{table}[!t]
 \caption{Robustness of the proposed method: All three detection models are trained by using images with straight text instances in the dataset Total-Text and then evaluated over images with curved text instances in the dataset Total-Text. Evaluations show that compared with MSR \cite{msr_2019_ijcai} and PSENet \cite{psenet}, our method is more robust and can train much better detection models by using straight text instances for detecting curved text instances.}
\centering
 \begin{tabular}{|c | c | c | c |} 
 \hline
 \textbf{Methods} & \textbf{Recall} & \textbf{Precision} & \textbf{F-score} \\ 
 \hline\hline
 MSR \cite{msr_2019_ijcai}                  & 50.1              & 68.8              & 58.3          \\
 PSENet \cite{psenet}                       & 51.3              & 72.3              & 60.0          \\
 \hline\hline
 \textbf{Ours}                              & \textbf{74.7}     & \textbf{77.3}     & \textbf{76.0} \\
 \hline
 \end{tabular}
 \label{tab:total_curve}
\end{table}

\begin{figure}[!t]
  \centering
  \includegraphics[width=\linewidth]{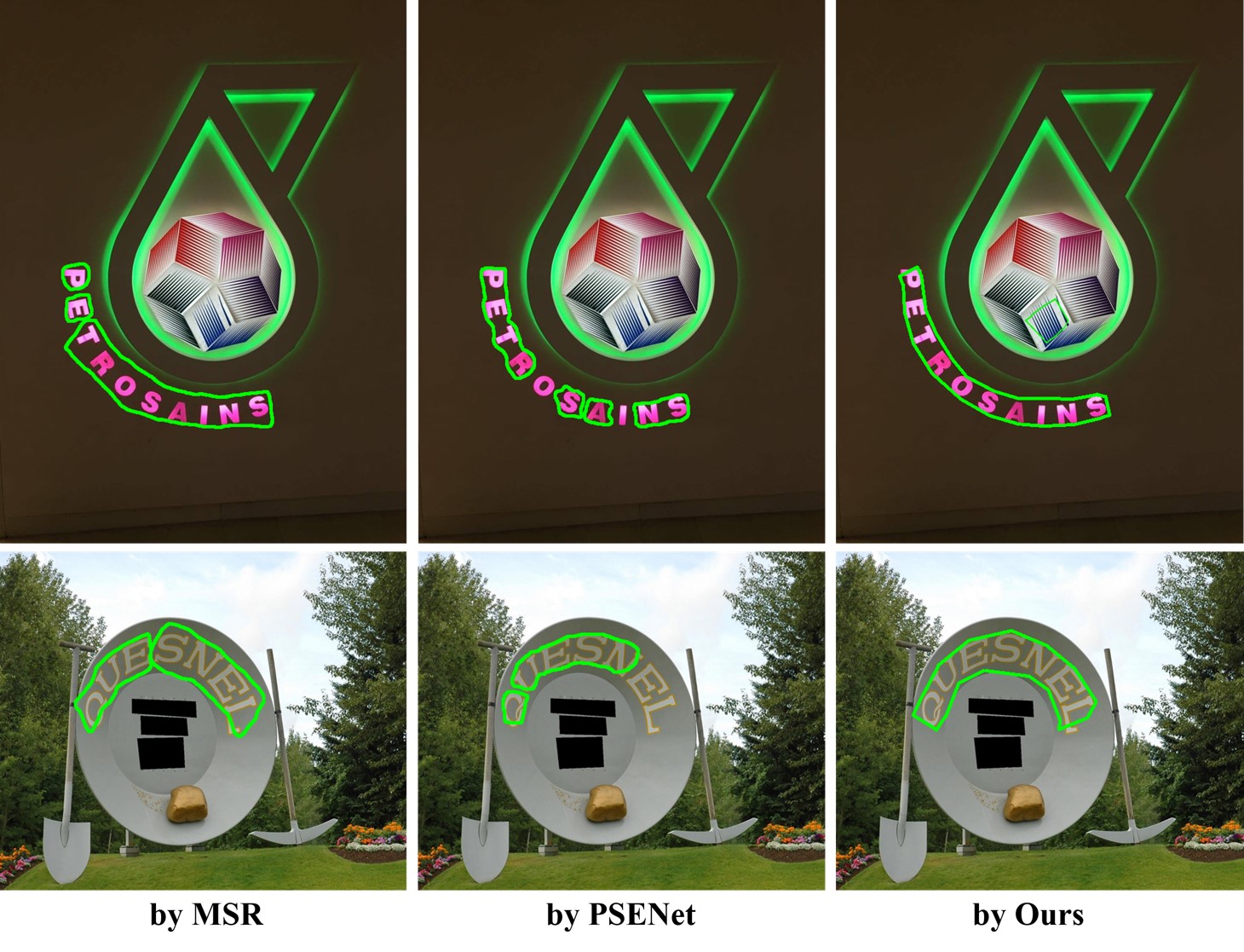}
\caption{Our proposed method is more tolerant to the variation in text line shapes. For detection models trained by using straight text instances in Total-Text, our model performs clearly much better than MSR and PSENet models while applied to the curved text instances in Total-Text.} 
\label{fig:curve_comp}
\end{figure}

\begin{figure}[t!]
  \centering
  \includegraphics[width=\linewidth]{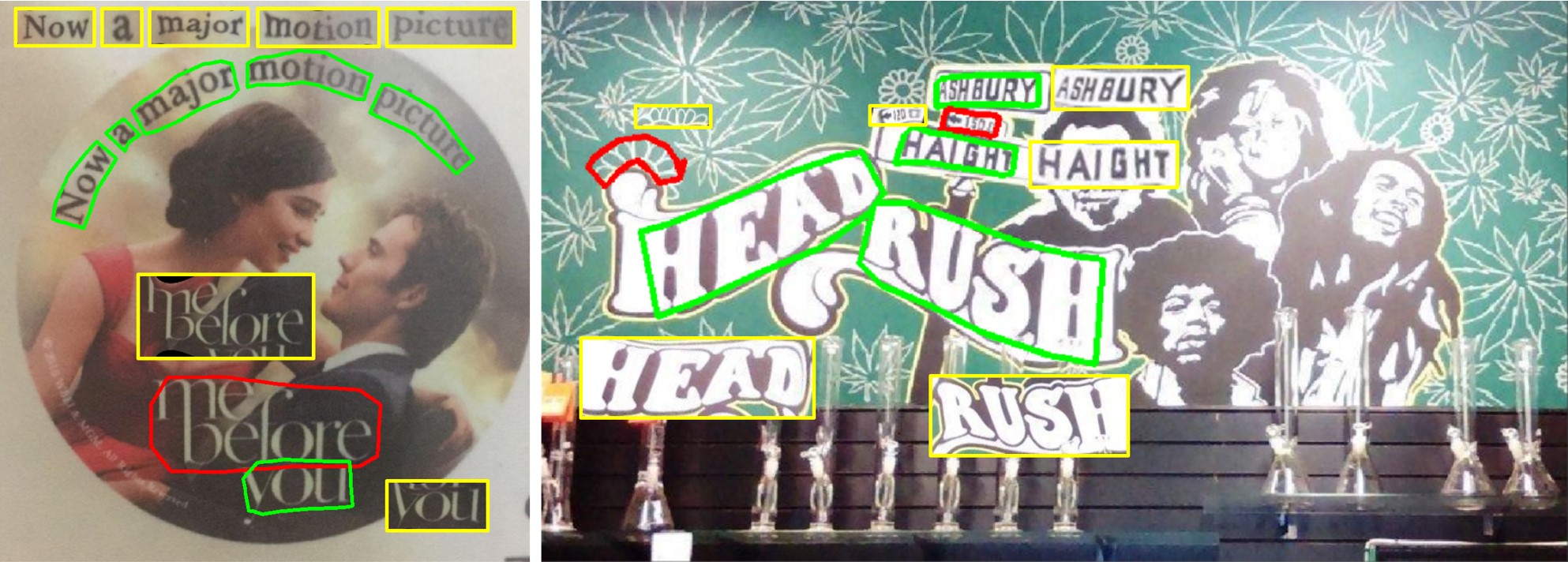}
\caption{Typical failure cases of the proposed method. Green boxes show the correct detection and red boxes show red boxes show the false detection. The rectified text instances are highlighted by yellow boxes.}
\label{fig:fail}
\end{figure}

\subsection{Discussion}
\subsubsection{Robustness} The proposed technique is tolerant to the rich variation in text lines shapes, lengths and orientations. We conduct a new experiment on Total-Text for verification. Specifically, we first generate a set of training images from the Total-Text training images that contain only straight text instances as annotated by quadrilateral bounding boxes. Similarly, we also generate a set of test images from the test images of Total-Text that contains only curved text instances as annotated by polygon bounding boxes.

Table \ref{tab:total_curve} and Fig. \ref{fig:curve_comp} show quantitative and qualitative results of our proposed method and two state-of-the-art methods while we train detection models by using the straight text instances and test over the curved text instances. As Table \ref{tab:total_curve} shows, the proposed methods outperforms MSR and PSENet by over 16\% in f-score. The large improvement can also be observed in Fig. \ref{fig:curve_comp}, where our proposed method can detect very curved text instances successfully whereas both MSR and PSENet fail. The much better robustness is largely attribute to the use of text keypoints that takes a bottom-up approach and so is not sensitive to the variation in text line shapes. As a comparison, MSR and PSENet tend to remember the text line shapes of training images and cannot handle images with different shapes well.

\subsubsection{Failure Cases} The proposed method still faces certain constraints under several specific scenarios. First, it may produce false detection if the background contains certain pattern that is similar to text strokes as shown in the first example in Fig. \ref{fig:fail}. Second, it could miss or produce false detection when multiple text lines are extremely close to each other as shown in the second example image in Fig. \ref{fig:fail}.

%-------------------------------------------------------------------------
\section{Conclusion}\label{sec:conclusion} 
This paper presents a novel scene text detection and rectification technique through keypoint detection. It formulates the scene text detection and rectification as a keypoint detection task and defines three types of text-specific keypoints. In addition, it designs a mask-guided multi-task network that simultaneously learns both of the character and the text instance features for detecting text keypoints and predicting keypoint links accurately. With the detected text keypoints and keypoint links, landmark points along the text boundary can be detected which can be exploited to locate and rectify text instances of arbitrary shapes reliably. Extensive experiments show that the proposed technique achieves superior detection and rectification performance.

\bibliographystyle{elsarticle-num}
\bibliography{mybibfile}

\end{document}